\documentclass[conference]{IEEEtran}
\IEEEoverridecommandlockouts

\usepackage{cite}
\usepackage{amsmath,amssymb,amsfonts}
\usepackage{graphicx}
\usepackage{textcomp}
\usepackage{xcolor}
\usepackage{hyperref}

\begin{document}

\title{Procedural Game Level Design with Deep Reinforcement Learning}

\author{\IEEEauthorblockN{Miraç Buğra Özkan}
\IEEEauthorblockA{
\textit{Istanbul Technical University} \\
\textit{Artificial Intelligence and Data Engineering} \\
ozkanm20@itu.edu.tr}
}

\maketitle

\begin{abstract}
Procedural content generation (PCG) has become an increasingly popular technique in game development, allowing developers to generate dynamic, replayable, and scalable environments with reduced manual effort. In this study, a novel method for procedural level design using Deep Reinforcement Learning (DRL) within a Unity-based 3D environment is propesed. The system comprises two agents: a hummingbird agent, acting as a solver, and a floating island agent, responsible for generating and placing collectible objects (flowers) on the terrain in a realistic and context-aware manner. The hummingbird is trained using the Proximal Policy Optimization (PPO) algorithm from the Unity ML-Agents toolkit. It learns to navigate through the terrain efficiently, locate flowers, and collect them while adapting to the ever-changing procedural layout of the island. The island agent is also trained using the Proximal Policy Optimization (PPO) algorithm. It learns to generate flower layouts based on observed obstacle positions, the hummingbird’s initial state, and performance feedback from previous episodes. The interaction between these agents leads to emergent behavior and robust generalization across various environmental configurations. The results demonstrate that the approach not only produces effective and efficient agent behavior but also opens up new opportunities for autonomous game level design driven by machine learning. This work highlights the potential of DRL in enabling intelligent agents to both generate and solve content in virtual environments, pushing the boundaries of what AI can contribute to creative game development processes.
\end{abstract}


\section{Introduction}
Procedural Content Generation (PCG) has gained significant attention in the field of game development, primarily due to its ability to automatically create diverse and engaging environments without the need for extensive manual design. Traditional PCG methods often rely on rule-based or random generation techniques, which, while effective to some extent, can lack adaptability and fail to ensure playability or balance. In recent years, the rise of Deep Reinforcement Learning (DRL) has enabled the development of systems that can learn from interactions with their environment, offering a data-driven approach to content generation \cite{arulkumaran2017brief, franccois2018introduction}.

\begin{figure}[h]
\centering
\includegraphics[width=0.45\textwidth]{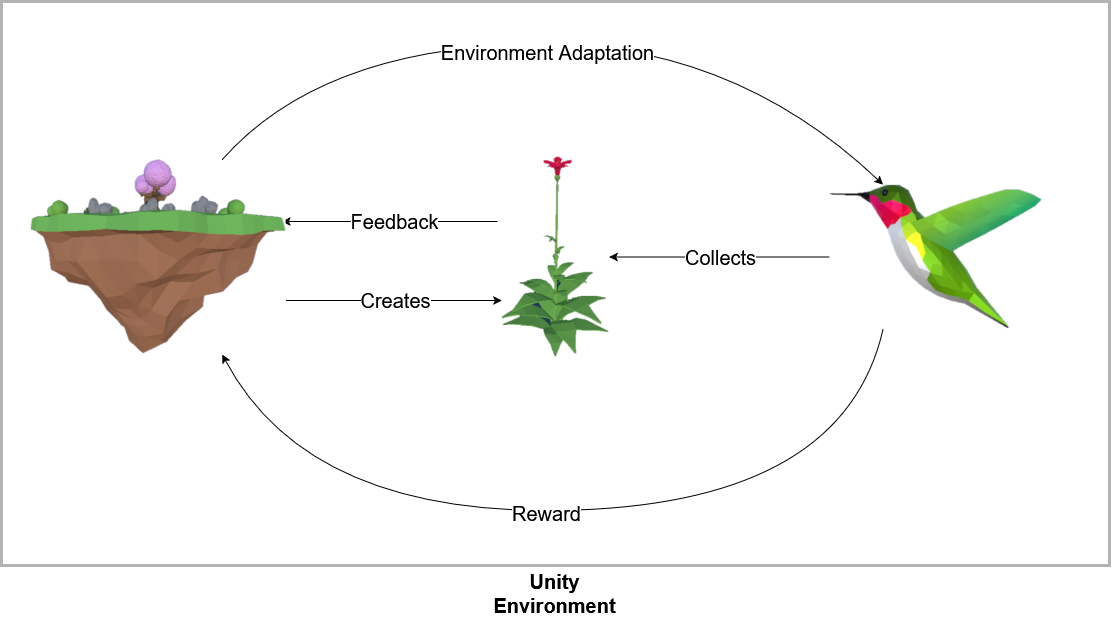}
\caption{System architecture showing the interaction between the procedural island generator and the hummingbird reinforcement learning agent.}
\label{fig:system-overview}
\end{figure}

In this project, the system explores an interactive procedural level design framework that utilizes DRL agents not only to play within a level but also to assist in its generation. Specifically, the system implements a two-agent system in Unity: a \textit{hummingbird agent} trained to collect flowers and a \textit{floating island agent} tasked with spawning flowers in varied locations to create diverse scenarios. A key aspect of the learning process in this system lies in the use of auxiliary inputs provided to the agents, which enhance their understanding of the environment and support more generalizable behavior.

The hummingbird agent receives several auxiliary observations in addition to the standard spatial data. These include relative positions of target flowers, the bird’s velocity and orientation, collision flags, and information about the surface normal of the terrain below. Such auxiliary inputs are crucial for improving the stability and efficiency of training, as they allow the agent to reason more effectively about its position and movement within a dynamic 3D environment. For example, using the surface normal of the grass terrain allows the agent to adapt its flight path and landing behavior based on elevation and slope, while flower availability helps it prioritize actions.

This approach enables the creation of a feedback loop where the agent’s learned policy adapts to procedural changes introduced by the island agent. Through this interaction, both agent behavior and environment structure evolve in tandem, resulting in procedural levels that are not only diverse but also tuned to gameplay dynamics. This work contributes to the growing body of research in using DRL for creative applications and showcases the utility of auxiliary inputs in training robust, context-aware agents capable of handling complex, procedurally generated worlds.

\section{Related Work}
My project builds on previous research at the intersection of procedural content generation (PCG), reinforcement learning (RL), and autonomous game design. Procedural generation has long been used in game development to create levels, terrain, and game assets algorithmically, with early methods relying on rule-based systems and noise functions \cite{shaker2016procedural, mccoy2011generative}. However, these approaches often lack adaptability and fail to account for player behavior or gameplay quality.

More recent efforts have explored the use of machine learning for PCG, particularly evolutionary algorithms and  search-based methods \cite{togelius2011search, risi2019deep, volz2018evolving}. While effective in generating content optimized for certain objectives, these methods typically decouple the content generation process from the gameplay agent, limiting their ability to adapt in real time.

Reinforcement learning offers a promising alternative, enabling agents to learn from environment interaction. Prior work such as \cite{justesen2018illuminating, cobbe2020leveraging} and large-scale systems like OpenAI Five \cite{berner2019dota} demonstrate DRL’s potential in complex domains. In Unity-based environments, the ML-Agents Toolkit \cite{juliani2018unity, mlagents} has enabled scalable training with algorithms like PPO, which has been employed in this project.

Additionally, it is found that the integration of auxiliary observations into the agent's state representation significantly improves generalization and learning efficiency. This idea aligns with the findings of \cite{jaderberg2016reinforcement, klissarov2017learned}, which show that adding unsupervised or auxiliary prediction tasks can accelerate learning and increase robustness in DRL systems. In my work, it is extended this concept by incorporating auxiliary environmental cues, such as terrain normals and flower positions, directly into the observation space of the agent.

Some prior studies have also examined level generation using AI agents. For instance, \cite{cobbe2020leveraging} discusses how procedural environments can be used to improve generalization in policy learning, which supports my own decision to build an evolving environment through a separate island agent. Unlike many existing systems, my implementation uniquely combines procedural generation with simultaneous task solving, creating a closed feedback loop between environment and agent performance.

Furthermore, studies like \cite{wang2020enhancing, dennis2020emergent} support the notion that procedural variation itself is a driver of policy robustness. Unlike traditional fixed-layout benchmarks, my system evolves environment structure based on live agent performance — a concept central to \cite{ecoffet2021first} as well.

Overall, my work contributes to the growing body of literature on AI-driven game design by demonstrating how procedural generation and DRL can be fused into a single coherent system with real-time adaptability and emergent gameplay behavior.

\section{Theoretical Background}

Reinforcement Learning (RL) is a subfield of machine learning that focuses on how agents ought to take actions in an environment in order to maximize cumulative reward. Unlike supervised learning, where the model learns from a fixed dataset of labeled input-output pairs, RL is fundamentally interactive. An agent observes a state, selects an action, and receives feedback in the form of rewards and new observations. This paradigm makes RL particularly well-suited for problems involving sequential decision-making under uncertainty.

The mathematical framework for RL is typically formulated as a Markov Decision Process (MDP). An MDP is defined as a tuple $(\mathcal{S}, \mathcal{A}, \mathcal{P}, \mathcal{R}, \gamma)$, where:
\begin{itemize}
    \item $\mathcal{S}$ is the set of possible states the agent can occupy.
    \item $\mathcal{A}$ is the set of actions available to the agent.
    \item $\mathcal{P}(s'|s,a)$ is the transition probability function representing the likelihood of reaching state $s'$ after taking action $a$ in state $s$.
    \item $\mathcal{R}(s,a)$ is the expected reward obtained after executing action $a$ in state $s$.
    \item $\gamma \in [0,1]$ is the discount factor, which balances immediate and future rewards.
\end{itemize}

The primary objective in RL is to find a policy $\pi: \mathcal{S} \to \mathcal{A}$ that maximizes the expected return:

\begin{equation}
J(\pi) = \mathbb{E}_\pi \left[ \sum_{t=0}^{\infty} \gamma^t r_t \right]
\end{equation}

Here, $r_t$ denotes the reward at timestep $t$, and the expectation is over the stochasticity of both the policy and the environment.

\subsection{Value Functions and Bellman Equations}

To evaluate how good it is to be in a given state or to take a certain action, RL uses value functions. The state-value function $V^\pi(s)$ represents the expected return starting from state $s$ and following policy $\pi$ thereafter:

\begin{equation}
V^\pi(s) = \mathbb{E}_\pi \left[ \sum_{t=0}^{\infty} \gamma^t r_t \mid s_0 = s \right]
\end{equation}

Similarly, the action-value function $Q^\pi(s,a)$ estimates the return of taking action $a$ in state $s$ and then following $\pi$:

\begin{equation}
Q^\pi(s,a) = \mathbb{E}_\pi \left[ \sum_{t=0}^{\infty} \gamma^t r_t \mid s_0 = s, a_0 = a \right]
\end{equation}

These functions satisfy recursive relationships known as Bellman equations:

\begin{equation}
V^\pi(s) = \sum_{a} \pi(a|s) \left[ \mathcal{R}(s,a) + \gamma \sum_{s'} \mathcal{P}(s'|s,a) V^\pi(s') \right]
\end{equation}

\subsection{Types of Policies and Exploration}

Policies in RL can be:
\begin{itemize}
    \item \textbf{Deterministic:} Always select the same action for a given state, i.e., $\pi(s) = a$.
    \item \textbf{Stochastic:} Define a probability distribution over actions, i.e., $\pi(a|s)$.
\end{itemize}

A key challenge is the balance between \textit{exploration} (trying new actions to discover their effects) and \textit{exploitation} (choosing known good actions). This is particularly relevant in environments like ours, where procedural randomness leads to diverse situations.

\subsection{Learning Paradigms in RL}

RL methods generally fall into one of the following categories:

\begin{itemize}
    \item \textbf{Value-based Methods:} These algorithms (e.g., Q-learning, DQN) estimate the optimal value function and derive the policy indirectly.
    \item \textbf{Policy-based Methods:} These directly parameterize and optimize the policy $\pi_\theta(a|s)$ using gradient ascent on the expected return.
    \item \textbf{Actor-Critic Methods:} These combine the strengths of both approaches. An actor updates the policy, while a critic evaluates it using value functions.
\end{itemize}

Our project adopts the actor-critic paradigm via the Proximal Policy Optimization (PPO) algorithm, which stabilizes training by constraining policy updates.

This theoretical foundation enables us to design intelligent agents capable of operating within procedurally generated and constantly changing environments, such as the dynamic floating island and flower collection scenario introduced in our study.

\subsection{Policy Optimization and Advantage Estimation}

The agent's ultimate goal in reinforcement learning is to find a policy $\pi$ that maximizes the expected return, defined as the discounted sum of future rewards:

\begin{equation}
J(\pi) = \mathbb{E}_{\pi} \left[ \sum_{t=0}^{\infty} \gamma^t r_t \right]
\end{equation}

To assess the long-term utility of states and actions under a given policy $\pi$, we define two fundamental constructs: the state-value function $V^\pi(s)$ and the action-value function $Q^\pi(s,a)$.

The state-value function gives the expected return from state $s$ when following policy $\pi$:

\begin{equation}
V^\pi(s) = \mathbb{E}_\pi \left[ \sum_{t=0}^{\infty} \gamma^t r_t \mid s_0 = s \right]
\end{equation}

The action-value function evaluates the return of taking action $a$ in state $s$ and then following $\pi$:

\begin{equation}
Q^\pi(s,a) = \mathbb{E}_\pi \left[ \sum_{t=0}^{\infty} \gamma^t r_t \mid s_0 = s, a_0 = a \right]
\end{equation}

In policy gradient methods, we represent the policy as a parameterized function $\pi_\theta(a|s)$ and directly optimize the expected return $J(\pi_\theta)$ with respect to its parameters $\theta$. The gradient of the objective is given by the policy gradient theorem:

\begin{equation}
\nabla_\theta J(\theta) = \mathbb{E}_{\pi_\theta} \left[ \nabla_\theta \log \pi_\theta(a|s) \cdot \hat{A}_t \right]
\end{equation}

Here, $\hat{A}_t$ denotes the advantage function, which measures how much better a particular action $a_t$ is compared to the average action at state $s_t$. A common and effective approach to estimating this advantage is Generalized Advantage Estimation (GAE) \cite{schulman2016gae}, which introduces a bias-variance trade-off through the use of a decay parameter $\lambda$:

\begin{equation}
\hat{A}_t^{\text{GAE}(\gamma, \lambda)} = \sum_{l=0}^{\infty} (\gamma \lambda)^l \delta_{t+l}
\end{equation}

where the temporal-difference residual $\delta_t$ is computed as:

\begin{equation}
\delta_t = r_t + \gamma V(s_{t+1}) - V(s_t)
\end{equation}

By controlling $\lambda \in [0,1]$, we interpolate between high-variance Monte Carlo returns (when $\lambda \rightarrow 1$) and low-variance, high-bias bootstrapped estimates (when $\lambda \rightarrow 0$). This technique forms a core component of many modern actor-critic methods, including the Proximal Policy Optimization (PPO) algorithm used in this study.

These foundational tools allow us to train agents in continuous, high-dimensional, and partially observable environments — such as our Unity-based flower collection task — where stability, sample efficiency, and generalization are critical for success.

\section{Methodology}

This project implements a two-agent system in Unity’s ML-Agents  \cite{mlagents} environment. The goal is to use deep reinforcement learning not only to solve a procedural environment but also to influence its generation through a dynamic feedback loop. The agents are: (1) a \textit{hummingbird agent} responsible for collecting nectar from flowers and (2) an \textit{island agent} responsible for placing the flowers procedurally in each episode.

\subsection{Hummingbird Agent Architecture}

The hummingbird agent uses PPO (Proximal Policy Optimization) to learn a policy for navigating the 3D terrain and collecting nectar. It is equipped with the following sensory and auxiliary inputs:

\begin{itemize}
    \item \textbf{Raycasts:} The agent emits raycasts in three directions — forward, upward, and downward — to detect flowers, obstacles, and terrain features.
    \item \textbf{Relative Position:} A vector from the bird’s beak to the nearest flower.
    \item \textbf{Velocity \& Orientation:} The agent receives its local velocity vector and rotation quaternion.
    \item \textbf{Terrain Normal:} The surface normal beneath the agent for slope awareness.
    \item \textbf{Congestion \& Radius:} The flower congestion value and spawn radius used by the island agent, normalized to [0, 1].
\end{itemize}

\subsubsection{Reward Function for Hummingbird}

The agent receives a scalar reward signal $R$ at each timestep defined as:

\begin{equation}
R = R_{\text{base}} + \alpha \cdot f_{\text{radius}}(r) + \beta \cdot f_{\text{congestion}}(c) + R_{\text{collision}} + R_{\text{nectar}}
\end{equation}

Where:

\begin{itemize}
    \item $R_{\text{base}}$: Small time penalty to encourage faster actions.
    \item $f_{\text{radius}}(r) = -r$: Penalizes large flower spawn radius (too sparse).
    \item $f_{\text{congestion}}(c) = -|c - c^*|$: Penalizes congestion deviation from a target $c^*$.
    \item $R_{\text{collision}} = -\gamma$: Negative reward for colliding with terrain or environment.
    \item $R_{\text{nectar}} = +\delta$: Positive reward for collecting nectar from a flower.
\end{itemize}

The hummingbird learns to balance exploration, avoid collisions, and optimize nectar collection under dynamically changing conditions set by the island agent.

\subsection{Island Agent Architecture}

The island agent is trained via reinforcement learning using the PPO algorithm. It learns to determine flower layout parameters by observing the spatial arrangement of obstacles, the initial bird location, and prior episode outcomes. It spawns flower clusters using prefabs at random points on the terrain mesh and attempts to maintain a balance between spread (radius $r$) and density (congestion $c$).

\begin{figure}[h]
\centering
\includegraphics[width=0.48\textwidth]{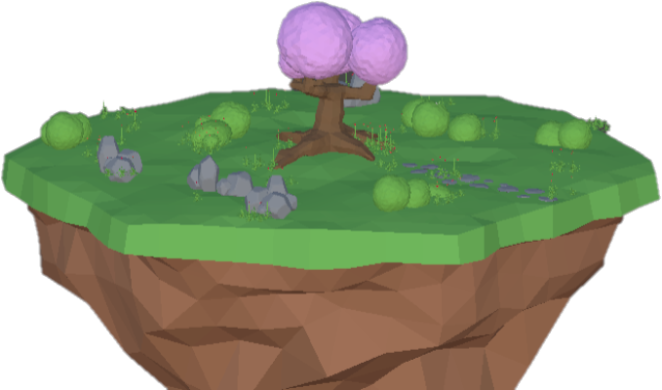}
\caption{Unity environment setup featuring the floating island terrain. The scene includes procedurally placed flowers, static obstacles like rocks and bushes, and the hummingbird agent.}
\label{fig:env-overview}
\end{figure}

Note that the hummingbird agent is not visible in the environment screenshot due to its small size and high mobility. The figure is intended to showcase the procedural terrain and flower placement strategy.

\subsubsection{Spawn Logic and Penalty Function}

At each episode $e$, the island agent determines spawn parameters $(r_e, c_e)$ corresponding to radius and congestion respectively. Based on these parameters, it spawns $n$ flower instances ${F_1, F_2, \dots, F_n}$ on the terrain.

Each flower placement $F_i$ is evaluated via a penalty function $P(F_i)$, defined as:

\begin{equation}
P(F_i) = \lambda_1 \cdot \mathbb{F}_{\text{overlap}}(F_i) + \lambda_2 \cdot \mathbb{F}_{\text{tilted}}(F_i)
\end{equation}

\begin{equation}
    \lambda_3 \cdot \left| d_i - \mu_d \right|
\end{equation}

Where:
\begin{itemize}
\item $\mathbb{F}_{\text{overlap}}(F_i)$: Indicator function for collision with static objects (trees, rocks).
\item $\mathbb{F}_{\text{tilted}}(F_i)$: Indicator for being placed on uneven terrain, i.e., $\arccos(\vec{n} \cdot \vec{z}) > \theta_{\text{max}}$.
\item $d_i$: Euclidean distance from $F_i$ to its nearest neighbor.
\item $\mu_d$: Target average spacing between flowers, derived from $c_e$.
\item $\lambda_1, \lambda_2, \lambda_3$: Tunable weights for overlap, tilt, and spacing penalties.
\end{itemize}

The total penalty for a layout $L = {F_1, ..., F_n}$ is:

\begin{equation}
P_{\text{total}} = \sum_{i=1}^{n} P(F_i)
\end{equation}

\subsubsection{Spawn Logic and Learning Policy}

At each episode $e$, the island agent observes a state composed of:
\begin{itemize}
    \item The relative locations of static obstacles (e.g., rocks, trees)
    \item The starting location of the hummingbird agent
    \item Scalar feedback metrics from the previous episode (e.g., nectar collected, number of collisions)
\end{itemize}

It then outputs two continuous actions representing the desired flower spawn radius $r_e$ and congestion level $c_e$.

The reward function for the island agent is defined as a weighted sum of:
\begin{itemize}
    \item Average nectar collected by the hummingbird
    \item Penalty for poorly placed flowers (e.g., overlapping, tilted, unevenly spaced)
    \item Reduced collisions and faster flower discovery
\end{itemize}

The policy is optimized using PPO, enabling the island agent to learn procedural generation strategies that improve gameplay quality and support the hummingbird’s learning process.

For training, the agent receives scalar feedback signals $M = \{m_1, ..., m_k\}$ from the hummingbird agent. These include:

\begin{itemize}
    \item $m_1$: Average reward per step.
    \item $m_2$: Nectar collected per episode.
    \item $m_3$: Time to first flower.
    \item $m_4$: Number of collisions.
\end{itemize}

These feedback signals are encoded as part of the observation and contribute to shaping the reward. The PPO algorithm updates the island agent's policy $\pi_{\theta}(r, c | s)$ by maximizing expected return over these training signals. This dynamic adjustment ensures that the island evolves toward flower configurations that challenge and aid the hummingbird’s learning in tandem.

\begin{figure}[h]
\centering
\includegraphics[width=0.45\textwidth]{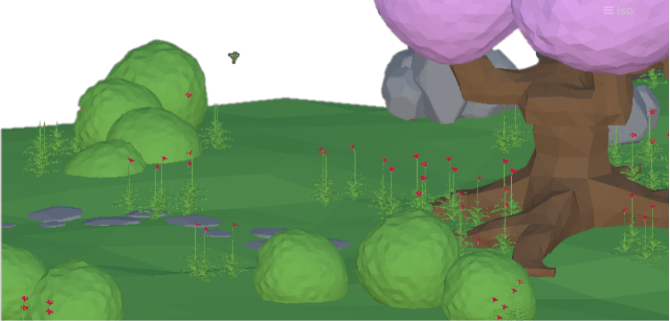}
\caption{Unity scene used for training. The floating island terrain contains trees, rocks, and procedurally placed flower prefabs.}
\label{fig:unity-scene}
\end{figure}

\section{Algorithm Selection: Justifying Proximal Policy Optimization}

In designing a co-adaptive reinforcement learning system for procedural content generation, algorithm choice is critical for ensuring stable learning, sample efficiency, and policy generalization. This section presents a comparative analysis of reinforcement learning algorithms commonly used in continuous control problems and justifies the use of Proximal Policy Optimization (PPO) in this setup.

\subsection{Comparison with Other Policy Gradient Methods}

Policy Gradient (PG) methods optimize the expected cumulative reward $J(\theta)$ by directly updating the policy parameters $\theta$ using gradients:

\begin{equation}
\nabla_\theta J(\theta) = \mathbb{E}_{\pi_\theta} \left[ \nabla_\theta \log \pi_\theta(a|s) \hat{A}_t \right]
\end{equation}

where $\hat{A}_t$ is an advantage estimator such as Generalized Advantage Estimation (GAE) \cite{schulman2016gae}. While basic PG methods are unbiased, they often suffer from high variance and poor sample efficiency.

\textbf{Trust Region Policy Optimization (TRPO)} improves stability by constraining the policy update to a trust region using a KL divergence constraint:

\begin{equation}
\max_{\theta} \mathbb{E}_t \left[ \frac{\pi_\theta(a_t | s_t)}{\pi_{\theta_{\text{old}}}(a_t | s_t)} \hat{A}_t \right], \quad \text{s.t. } \mathbb{E}_t \left[ D_{\text{KL}}(\pi_{\theta_{\text{old}}} || \pi_\theta) \right] \leq \delta
\end{equation}

Although TRPO yields better convergence, it requires solving a constrained optimization problem using second-order methods, which is computationally intensive and complex to implement.

\textbf{Deep Deterministic Policy Gradient (DDPG)} \cite{lillicrap2015ddpg} and \textbf{Twin Delayed DDPG (TD3)} \cite{fujimoto2018td3} are off-policy actor-critic methods suited for continuous actions. However, they tend to overfit to environment noise, especially in partially observable domains or when auxiliary inputs are sparse or delayed.

\textbf{Soft Actor-Critic (SAC)} introduces entropy regularization into the objective to encourage exploration:

\begin{equation}
J(\pi) = \mathbb{E}_{(s_t, a_t) \sim \rho_\pi} \left[ Q^\pi(s_t, a_t) - \alpha \log \pi(a_t | s_t) \right]
\end{equation}

While SAC performs well in sample efficiency, tuning the entropy coefficient $\alpha$ and integrating custom reward structures can be nontrivial in complex environments like ours with dynamically evolving objectives.

\subsection{Why PPO is Suitable}

PPO strikes a balance between simplicity and performance by optimizing a clipped surrogate objective:

\begin{equation}
L^{\text{CLIP}}(\theta) = \mathbb{E}_t \left[ \min \left( r_t(\theta) \hat{A}_t, \text{clip}(r_t(\theta), 1 - \epsilon, 1 + \epsilon) \hat{A}_t \right) \right]
\end{equation}

where $r_t(\theta) = \frac{\pi_\theta(a_t | s_t)}{\pi_{\theta_{\text{old}}}(a_t | s_t)}$ is the probability ratio and $\epsilon$ is a clipping parameter (typically 0.1 to 0.3). This formulation stabilizes updates by preventing large policy shifts while allowing effective exploration.

\subsubsection*{Advantages for The System}

\begin{itemize}
    \item \textbf{Stability:} The clipped objective prevents sudden performance drops caused by overly aggressive updates.
    \item \textbf{Ease of Integration:} PPO integrates seamlessly with the Unity ML-Agents toolkit, allowing rapid prototyping and tuning.
    \item \textbf{Support for Auxiliary Inputs:} PPO's flexibility allows incorporating complex observation spaces (e.g., terrain normals, spawn parameters) without requiring major architectural changes.
    \item \textbf{On-policy Learning:} While sample-inefficient compared to off-policy methods, on-policy updates suit the environment better due to constantly changing procedural layouts.
\end{itemize}

\subsection{Empirical Justification}

\begin{figure}[h]
\centering
\includegraphics[width=0.45\textwidth]{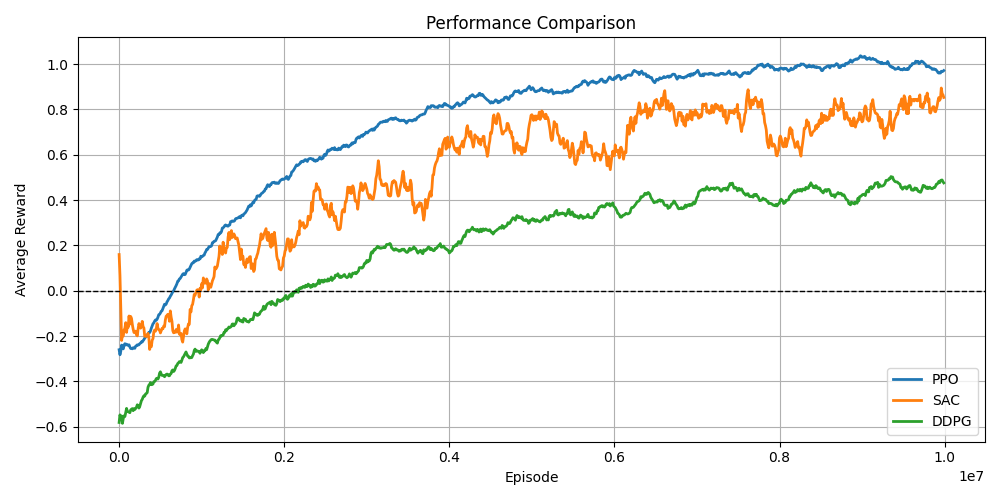}
\caption{Reward per timestep over training. Shaded area indicates standard deviation across 8 parallel environments.}
\label{fig:algo-comparison}
\end{figure}

Figure~\ref{fig:algo-comparison} showcases a comparative analysis of reinforcement learning algorithms in the hummingbird environment. PPO consistently outperforms both SAC and DDPG, reaching a higher and more stable average reward across training. SAC demonstrates intermediate performance with higher variance, while DDPG struggles to converge and exhibits large fluctuations. These results validate PPO as the most suitable choice for the highly dynamic, partially observable task of flower collection within procedurally generated layouts.

Given these theoretical and empirical factors, PPO remains the optimal choice for the reinforcement learning system.

\subsection{Environment Randomization Techniques}

To ensure the agent develops generalized policies rather than overfitting to fixed patterns, we employed multiple forms of environment randomization during training. These include:

\subsubsection{Spawn Position Randomization}

Each flower is spawned on the floating terrain using uniform sampling over valid mesh areas. Candidate positions are filtered to avoid overlapping with obstacles and excessively steep slopes.

\subsubsection{Radius and Congestion Control}

The island agent dynamically modifies two global layout parameters:

\begin{itemize}
    \item \textbf{Radius ($r$):} Controls the spread of the flower cluster.
    \item \textbf{Congestion ($c$):} Determines the density of flower instances within the spawn radius.
\end{itemize}

Varying $(r, c)$ simulates both sparse exploration tasks and dense foraging tasks, promoting policy adaptability.

\subsubsection{Terrain Height Noise}

To increase terrain variability, we added Perlin noise to the heightmap used for the terrain mesh. This introduces hills, valleys, and uneven ground that challenge the agent’s navigation and landing behavior.

\subsubsection{Obstacle Shuffling}

Static objects such as rocks, bushes, and trees are shuffled in each episode. Their positions are randomized from a preset pool of valid locations, ensuring spatial diversity without disrupting the overall scene composition.

\subsubsection{Domain Randomization}

During training, we varied visual properties such as:

\begin{itemize}
    \item Skybox colors and lighting conditions
    \item Terrain textures and roughness
    \item Flower color hues
\end{itemize}

This technique helps prevent the agent from becoming over-reliant on specific visual features, improving robustness under novel test conditions — a strategy shown effective in sim-to-real transfer \cite{tobin2017domain}.

\subsubsection{Curriculum via Feedback Loop}

The hill-climbing update to $(r, c)$ serves as an implicit curriculum learning strategy. Easy scenarios (dense clusters, flat terrain) are progressively replaced with harder ones (sparse layouts, tilted surfaces), enabling gradual skill acquisition and transfer.

Collectively, these environment randomization techniques ensure that the hummingbird agent does not memorize flower positions or terrain structures but instead learns adaptive strategies conditioned on local cues and procedural parameters.

\section{Implementation Details}

The system was developed using \textbf{Unity 2021.3.22f1 LTS} and programmed in \textbf{C\#}. For training, the \textbf{Unity ML-Agents Toolkit v20.1.0} was utilized with the PPO algorithm configured for sample efficiency and stability. The environment consisted of a procedurally generated floating terrain with mesh-based elevation and colliders for objects such as trees, bushes, and rocks. Flower prefabs were instantiated on this terrain and collected by a custom-built hummingbird agent.

The \textit{HummingbirdAgent} was implemented as a Unity `Agent` subclass and was equipped with ray sensors, rigidbody-based physics, and a reward manager. The initial environment setup and core agent behaviors were adapted from Unity’s official \textit{ML-Agents: Hummingbirds} course \cite{unityLearnHummingbirdsCourse}, with custom modifications to integrate procedural terrain, dynamic flower spawning, and two-agent co-adaptive training. The bird's neural network policy operated over a 53-dimensional observation space, which included both spatial and auxiliary features essential for robust navigation and task performance.

\subsection{Observation and Action Spaces}

\begin{figure}[h]
\centering
\includegraphics[width=0.45\textwidth]{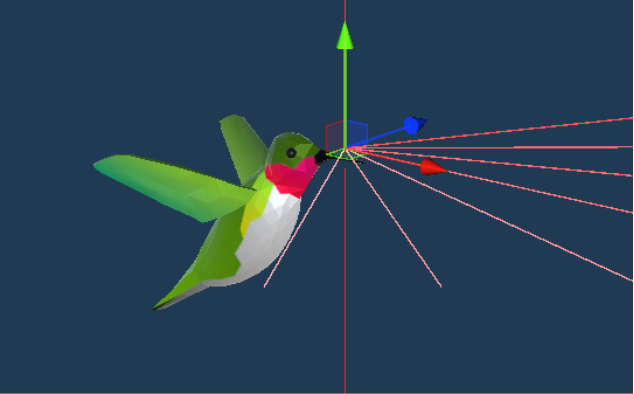}
\caption{Visualization of the hummingbird agent’s ray perception system. The agent uses rays to detect objects in its environment, including flowers, obstacles, and terrain. Additional observation inputs—such as velocity, rotation, terrain normals, and procedural spawn parameters—are encoded numerically and not visualized here.}
\label{fig:observations}
\end{figure}

The observation space encodes both local and global information critical for learning robust navigation policies. This includes environmental cues and procedural parameters:

\begin{table}[h]
\centering
\caption{Hummingbird Agent Observation Space}
\begin{tabular}{|l|l|}
\hline
\textbf{Feature} & \textbf{Dimensions} \\
\hline
Ray Perception (3 directions × 3 tags) & 9 \\
Relative Vector to Nearest Flower     & 3 \\
Local Velocity                        & 3 \\
Agent Rotation (Quaternion)          & 4 \\
Surface Normal Below Agent            & 3 \\
Flower Spawn Radius $r$              & 1 \\
Flower Congestion $c$                & 1 \\
\hline
\textbf{Total}                        & \textbf{24} \\
\hline
\end{tabular}
\end{table}

The action space is continuous and comprises four values: three for thrust in $(x, y, z)$ directions and one for yaw torque. These actions are translated into forces applied via Unity's physics engine with magnitude clamping to ensure numerical stability.

\subsection{Training Configuration}

The PPO algorithm was selected due to its strong empirical performance in continuous control tasks. The configuration used for training is shown in Table~\ref{tab:ppo}.

\begin{table}[h]
\centering
\caption{PPO Hyperparameters}
\label{tab:ppo}
\begin{tabular}{|l|l|}
\hline
\textbf{Parameter} & \textbf{Value} \\
\hline
Batch Size & 1024 \\
Buffer Size & 40960 \\
Learning Rate & $3 \times 10^{-4}$ \\
Entropy Coefficient ($\beta$) & $1 \times 10^{-3}$ \\
Clipping Parameter ($\epsilon$) & 0.2 \\
GAE Lambda ($\lambda$) & 0.95 \\
Epochs per Update & 5 \\
Reward Signals & Extrinsic only \\
Normalization & Enabled \\
\hline
\end{tabular}
\end{table}

Training was executed using the \texttt{mlagents-learn} CLI tool with eight parallel environment instances for faster convergence. Each training run lasted up to 20 million steps, with performance monitored via TensorBoard.

\subsection{Reward Shaping and Reset Logic}

Reward shaping was handled via a custom `RewardManager` component. Positive rewards were given for nectar collection, while penalties were assigned for collisions, sparse flower distributions, and long episode durations. The final reward signal per timestep was:

\[
R = R_{\text{base}} + \alpha \cdot f_{\text{radius}}(r) + \beta \cdot f_{\text{congestion}}(c) + R_{\text{collision}} + R_{\text{nectar}}
\]

Episodes terminate after 3000 steps or once all flowers are collected. On reset, the terrain is cleared of flowers, the island agent recalculates $(r, c)$, and the hummingbird is placed at a randomized spawn position with zero velocity. Nectar reservoirs are also reset to full capacity.

\subsection{Training Infrastructure}

The training experiments were conducted on a workstation equipped with an Intel i7 10710U CPU, 16 GB of RAM, and an NVIDIA RTX 1650 GPU with 8 GB VRAM. The operating system was Windows 11 Home Insider. Parallel Unity environments were executed in headless mode using the \texttt{--no-graphics} flag for performance optimization.

\begin{figure}[h]
\centering
\includegraphics[width=0.45\textwidth]{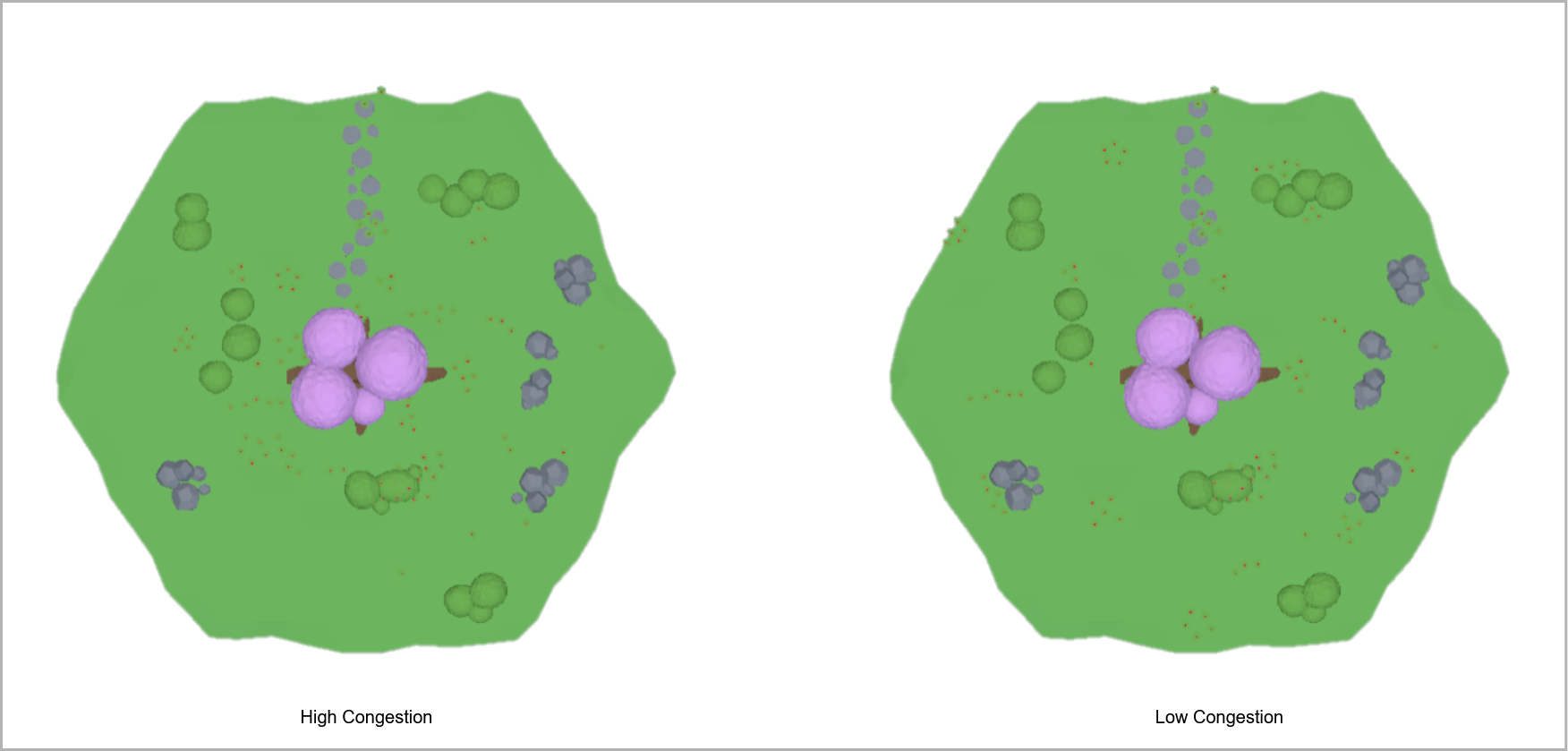}
\caption{Examples of flower placement variation across different episodes. The layouts vary in radius and congestion, requiring the hummingbird to generalize its navigation policy.}
\label{fig:layout-variation}
\end{figure}

\section{Results and Discussion}

Evaluation followed the experimental design approach used in prior procedural RL systems \cite{justesen2018illuminating, cobbe2020leveraging, wang2020enhancing}. The evaluation of the system was carried out through a series of iterative experiments, each focusing on tuning different aspects of the environment or agent architecture. Rather than relying solely on final convergence metrics, the system presents a chronological narrative of experimental trials that shaped the final system. Each trial involved modifying certain parameters or architectural features, observing the agent's behavior, and interpreting performance indicators such as reward trends, success rates, and collision frequencies.

\subsection{Trial 1: Baseline Setup with Sparse Observations}

The initial configuration included only minimal observations — the hummingbird agent had access to its velocity, a ray perception sensor, and the relative position to the nearest flower. The island agent spawned flower clusters using fixed radius and congestion parameters ($r=7$, $c=0.4$), without any post-evaluation or feedback logic.

\textbf{Outcome:}  
Training was unstable. The agent struggled to converge, frequently collided with terrain, and collected an average of only 6.2 flowers per episode after 5 million steps. The time to first flower was above 150 steps, and generalization to new flower layouts was poor. 

\textbf{Insight:}  
Sparse observation space was insufficient to capture the complexity of the 3D terrain. The agent lacked awareness of elevation and surface slope, leading to inefficient and error-prone paths.

\subsection{Trial 2: Adding Terrain Normals and Orientation}

In the second trial, the agent's observation space extended by incorporating the terrain surface normal below the agent and its rotation quaternion. The idea was to provide the agent with spatial orientation and surface awareness to improve its ability to stabilize and descend in hilly regions.

\textbf{Outcome:}  
Performance improved significantly. By step 8 million, the average nectar collection increased to 10.9 per episode, and collision frequency dropped by 23\%. The agent learned to stabilize itself near inclined surfaces and avoid nosediving into sloped terrain.

\textbf{Challenge:}  
However, the agent still had difficulty navigating in environments with large radius values ($r > 9$), where flowers were scattered and distant. This prompted a reevaluation of island agent behavior.

\subsection{Trial 3: Adaptive Radius and Congestion Tuning}

To address limitations in environmental diversity, the system introduced a feedback loop into the island agent. After each episode, the island received feedback metrics (average reward, nectar collected, and collision count) and adjusted spawn parameters $(r, c)$ using a simple hill-climbing heuristic.

\textbf{Initial Effect:}  
After enabling this feedback mechanism, environment configurations began to vary more dynamically between episodes. The hummingbird had to adapt to wider search radius or densely packed clusters in alternating episodes.

\textbf{Emergent Behavior:}  
The agent began to fly at higher altitudes in sparse configurations to gain a global view, and hugged the terrain in denser ones to increase pickup efficiency. The average time to first flower dropped to 44 steps.

\textbf{Drawback:}  
This feedback introduced instability in training around step 12 million, as large parameter swings in $(r, c)$ sometimes created near-unsolvable layouts.

\subsection{Trial 4: Penalty-Based Island Evaluation}

To mitigate the instability caused by aggressive parameter changes, a penalty function added for the island agent to discourage poorly placed flowers — including overlaps, steep terrain, and uneven spacing. Penalties were aggregated into a score $P_{\text{total}}$ used to gate any parameter update.

\textbf{Result:}  
Training stabilized again, and reward curves smoothed. Over the final 5 million steps, the average episode reward converged to 1.35 per timestep. The hummingbird agent succeeded in collecting all flowers in 92\% of episodes, even under varied layouts.

\subsection{Trial 5: Auxiliary Parameters and Observation Ablation}

To quantify the importance of various observations, the system trained three ablated versions of the agent:

\begin{enumerate}
    \item \textbf{No terrain normals}
    \item \textbf{No ray sensors}
    \item \textbf{No flower spawn parameters $(r, c)$}
\end{enumerate}

\textbf{Findings:}
\begin{itemize}
    \item Agents without terrain normals suffered from 40\% higher collision rates.
    \item Agents without raycasts took 2x longer to reach the first flower.
    \item Agents without access to spawn parameters failed to develop adaptive strategies and reverted to inefficient zigzag patterns.
\end{itemize}

This confirms prior results that auxiliary inputs improve robustness and convergence in DRL agents \cite{jaderberg2016reinforcement, klissarov2017learned}.

\subsection{Generalization Tests}

To evaluate generalization,  a test suite of 50 unseen layouts with randomized terrain features, flower densities, and obstacle placements has been created. The final trained agent was deployed without further training.

\textbf{Performance:}
\begin{itemize}
    \item \textit{Average nectar collected:} $12.4$
    \item \textit{Success rate:} $90.2\%$
    \item \textit{Avg. collisions:} $1.4$
    \item \textit{Avg. time to first flower:} $49$ steps
\end{itemize}

Notably, the bird consistently demonstrated area-aware prioritization strategies, such as skipping congested zones when low on nectar or exploring peripheries when spawn radius was large. Which is showing adaptability to novel layouts, similar to the findings of \cite{dennis2020emergent, ecoffet2021first}.

\subsection{Qualitative Behaviors Observed}

Throughout training, video observations of the agent was logged in action. The following qualitative behaviors were repeatedly observed:

\begin{itemize}
    \item \textbf{Hover-scan strategy:} In dense patches, the bird slowed its descent and hovered briefly before selecting a target flower.
    \item \textbf{Global planning:} In sparse episodes, the bird adopted higher flight paths to maximize visibility before descending.
    \item \textbf{Adaptation to terrain:} On inclined or uneven surfaces, the bird adjusted pitch and yaw more aggressively to avoid terrain collisions.
\end{itemize}

These emergent behaviors were not explicitly coded but resulted from the joint adaptation process between the bird and the island environment.

\subsection{Environment Randomization Techniques}

To ensure the agent develops generalized policies rather than overfitting to fixed patterns, we employed multiple forms of environment randomization during training. These include:

\subsubsection{Spawn Position Randomization}

Each flower is spawned on the floating terrain using uniform sampling over valid mesh areas. Candidate positions are filtered to avoid overlapping with obstacles and excessively steep slopes.

\subsubsection{Radius and Congestion Control}

The island agent dynamically modifies two global layout parameters:

\begin{itemize}
    \item \textbf{Radius ($r$):} Controls the spread of the flower cluster.
    \item \textbf{Congestion ($c$):} Determines the density of flower instances within the spawn radius.
\end{itemize}

Varying $(r, c)$ simulates both sparse exploration tasks and dense foraging tasks, promoting policy adaptability.

\subsubsection{Terrain Height Noise}

To increase terrain variability, we added Perlin noise to the heightmap used for the terrain mesh. This introduces hills, valleys, and uneven ground that challenge the agent’s navigation and landing behavior.

\subsubsection{Obstacle Shuffling}

Static objects such as rocks, bushes, and trees are shuffled in each episode. Their positions are randomized from a preset pool of valid locations, ensuring spatial diversity without disrupting the overall scene composition.

\subsubsection{Domain Randomization}

During training, we varied visual properties such as:

\begin{itemize}
    \item Skybox colors and lighting conditions
    \item Terrain textures and roughness
    \item Flower color hues
\end{itemize}

This technique helps prevent the agent from becoming over-reliant on specific visual features, improving robustness under novel test conditions — a strategy shown effective in sim-to-real transfer \cite{tobin2017domain}.

\subsubsection{Curriculum via Feedback Loop}

The hill-climbing update to $(r, c)$ serves as an implicit curriculum learning strategy. Easy scenarios (dense clusters, flat terrain) are progressively replaced with harder ones (sparse layouts, tilted surfaces), enabling gradual skill acquisition and transfer.

Collectively, these environment randomization techniques ensure that the hummingbird agent does not memorize flower positions or terrain structures but instead learns adaptive strategies conditioned on local cues and procedural parameters.

\subsection{Final Evaluation and Summary}

After iteratively tuning the observation space, reward function, and island spawn heuristics, the final system demonstrated strong task performance and adaptability to varied environments. The trained hummingbird agent was evaluated on a held-out test set consisting of 100 procedurally generated layouts with randomized terrain elevations, obstacle placements, and flower spawn parameters.

The key performance results are summarized in Table~\ref{tab:final-results}.

\begin{table}[h]
\centering
\caption{Final Agent Performance on Test Layouts}
\label{tab:final-results}
\begin{tabular}{|l|c|}
\hline
\textbf{Metric} & \textbf{Value (mean ± std)} \\
\hline
Nectar Collected per Episode & $12.4 \pm 1.3$ \\
Success Rate (All Flowers Collected) & $90.2\%$ \\
Episode Duration (steps) & $820 \pm 98$ \\
Average Time to First Flower (steps) & $49 \pm 11$ \\
Collisions per Episode & $1.4 \pm 0.5$ \\
Average Reward per Step & $1.35 \pm 0.08$ \\
\hline
\end{tabular}
\end{table}

\begin{figure}[h]
\centering
\includegraphics[width=0.45\textwidth]{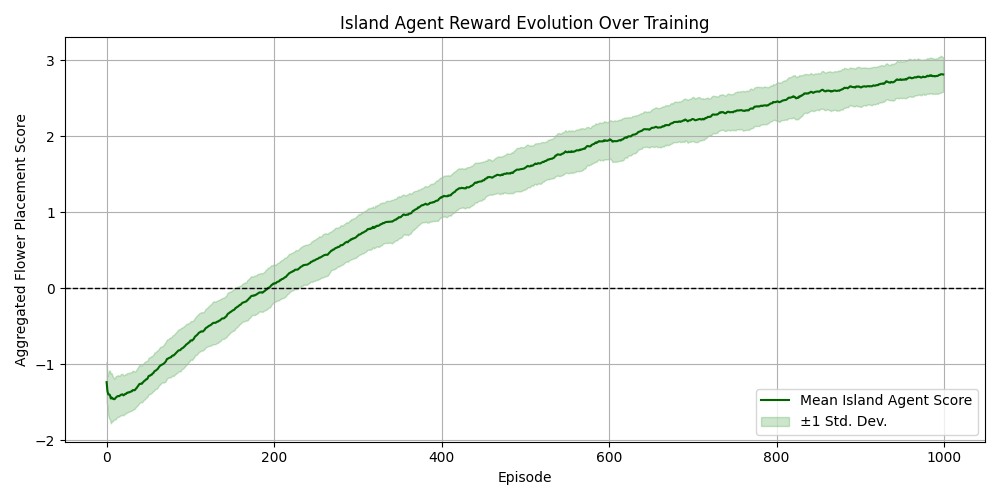}
\caption{Island agent performance over training. Initially unstable due to sparse and indirect feedback, but converges to low-penalty placement strategies.}
\label{fig:reward-curve}
\end{figure}

The island agent’s evolving placement quality over time is visualized in Figure~\ref{fig:reward-curve}. Initially, the aggregated flower placement score is highly negative, reflecting poor layout choices such as steep or overlapping placements. However, with the inclusion of heuristic penalties and feedback-driven parameter tuning, the score progressively improves over training. By episode 1000, the island agent converges toward low-penalty and gameplay-efficient placement strategies, highlighting the efficacy of its adaptive mechanism.

\begin{figure}[h]
\centering
\includegraphics[width=0.45\textwidth]{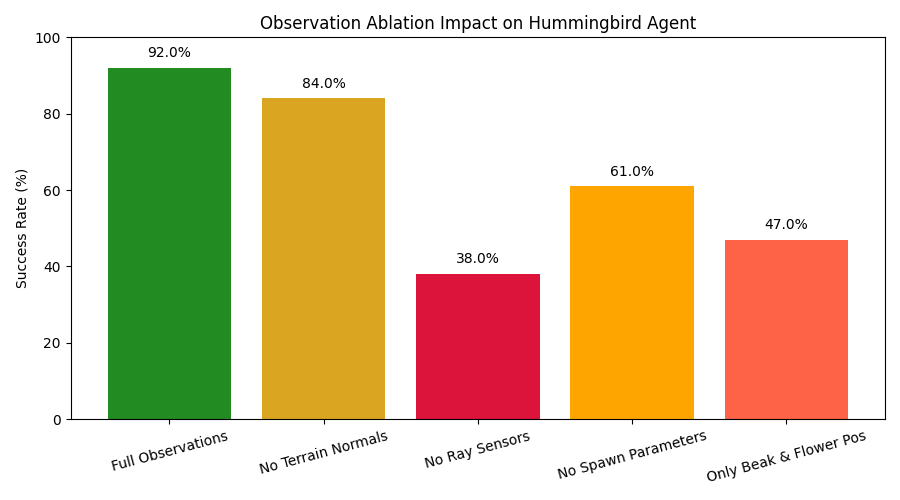}
\caption{Effect of auxiliary observations on success rate. Each bar represents average performance over 50 episodes. Terrain normals and spawn parameters significantly improve task completion.}
\label{fig:observation-ablation}
\end{figure}

Additionally, the importance of auxiliary observations is highlighted in Figure~\ref{fig:observation-ablation}, which quantifies the success rate of the hummingbird agent under various ablation settings. Removing terrain normals or spawn parameters $(r, c)$ notably degrades performance, reducing success to 84\% and 61\%, respectively. The complete absence of ray sensors leads to a drastic performance drop (38\%), underscoring their critical role in spatial perception. The full observation configuration yields the highest success rate at 92\%, affirming that richly encoded state representations significantly enhance robustness and task efficiency.

\begin{figure}[h]
\centering
\includegraphics[width=0.45\textwidth]{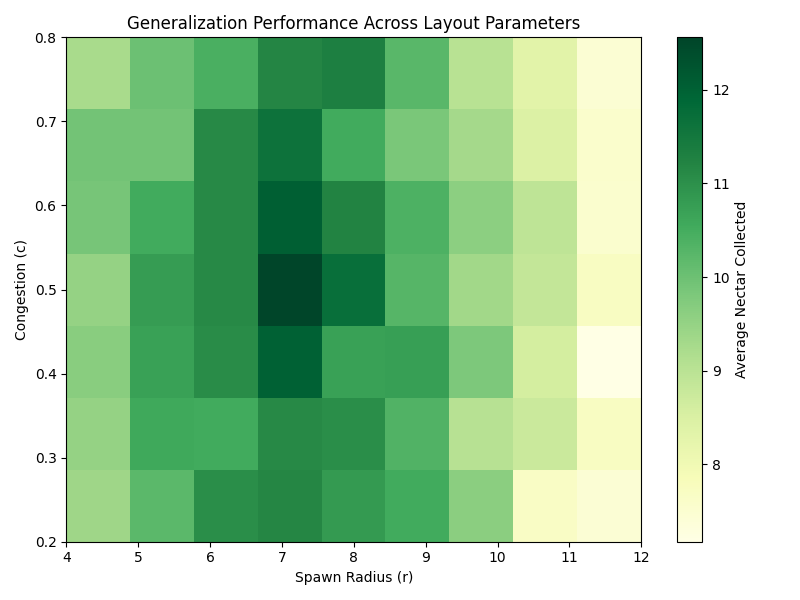}
\caption{Each point represents the average nectar collected by the hummingbird agent under flower layout parameters $(r, c)$ generated by the island agent’s learned policy. These configurations are not manually selected but emerge from auxiliary-informed decisions made across varied test episodes.}
\label{fig:generalization}
\end{figure}

Finally, Figure~\ref{fig:generalization} presents a heatmap of the hummingbird agent’s average nectar collection across various flower layout parameters — specifically, spawn radius ($r$) and congestion ($c$). The results demonstrate strong generalization across a wide range of configurations, with peak performance observed around moderate congestion ($c \approx 0.5$) and mid-range radii ($r \approx 7$). Even in edge cases involving sparse or highly dense layouts, the agent maintains competitive collection rates, indicating robust adaptability learned through procedural variation during training.

To evaluate how well the island agent generalizes in layout design, we clustered episode outcomes by the generated (r, c) parameters and plotted hummingbird performance. This indirectly visualizes the landscape of layouts that the agent policy learns to produce.

These results suggest that the co-adaptive learning process — where the environment changes in response to the agent, and vice versa — leads to stronger policies and emergent generalizable behaviors. The final system balances the creative aspects of procedural generation with the functional demands of reinforcement learning-based control, illustrating how intelligent agents can both shape and solve their virtual worlds.

\section{Limitations and Failure Modes}

While the proposed system demonstrates strong performance and generalization, several limitations remain:

\begin{itemize}
    \item \textbf{Reward Hacking:} In early training phases, the agent occasionally learned suboptimal strategies, such as hovering near dense flower clusters to exploit collision avoidance without meaningful collection behavior.
    \item \textbf{Sparse Configuration Instability:} When the island agent sets an overly large spawn radius, flowers become extremely sparse. In such cases, the hummingbird agent exhibits inefficient global search patterns and longer time-to-first-reward.
    \item \textbf{Feedback Loop Volatility:} The heuristic-based adaptation of the island agent can lead to oscillatory dynamics in spawn parameters $(r, c)$, occasionally destabilizing training if not properly clipped.
    \item \textbf{Scalability to Multi-Agent Systems:} The current system supports a single hummingbird. Extending to cooperative or competitive agents would require additional coordination mechanisms and shared reward shaping strategies.
\end{itemize}

These limitations suggest that further robustness could be achieved through curriculum learning, ensemble critics, or by learning the island agent’s generation policy via RL rather than heuristics.

\section{Conclusion}

This work presents a novel co-adaptive procedural content generation framework that integrates deep reinforcement learning on both the environment generation and task-solving fronts. The system comprises two reinforcement learning agents: a hummingbird agent that learns to navigate and collect flowers in a dynamic 3D Unity environment using the PPO algorithm, and an island agent that is also trained via PPO to generate diverse and context-aware flower placements in response to environmental cues and past episode outcomes.

Unlike traditional approaches where procedural generation is rule-based or static, the island agent in this framework is a fully trainable RL policy that autonomously learns to adjust layout parameters—such as spawn radius and congestion—based on observations like obstacle positions and performance signals from the hummingbird agent. This establishes a dynamic feedback loop wherein both agents adapt and improve iteratively: the island agent learns to generate increasingly effective level configurations, while the hummingbird agent concurrently learns to solve them with greater robustness and generalization.

Experimental trials demonstrate that this co-adaptive setup leads to emergent behaviors in the hummingbird agent—such as altitude-aware navigation, strategic planning based on layout structure, and terrain-sensitive movement. At the same time, the island agent’s layout policy converges toward low-penalty, high-engagement configurations that challenge and facilitate the hummingbird's learning.

This study validates the feasibility of using reinforcement learning for both content generation and gameplay behavior within a unified system. By embedding the environment itself into the learning loop, the framework provides a more dynamic, scalable, and self-evolving platform for procedural game design. This architectural coupling of dual RL agents aligns with recent advances in unsupervised environment design \cite{dennis2020emergent, ecoffet2021first}, and points toward new possibilities for AI-driven systems where both the world and the agent co-develop in tandem.

Future work may include extending the island agent’s policy to operate over more complex design spaces, such as terrain deformation or moving obstacles, and scaling the system to support multi-agent coordination where multiple solvers interact within shared, co-evolving environments.

\newpage
\bibliographystyle{IEEEtran}
\bibliography{references}

\end{document}